\documentclass[final,1p,times]{elsarticle}

\usepackage{amssymb}

\usepackage{graphicx}

\usepackage{bm}

\usepackage{amsfonts}
\usepackage{amsmath}

\usepackage{float}

\usepackage{multirow}
\usepackage{dcolumn}
\usepackage{xcolor}
\usepackage{boldline}
\usepackage[flushleft]{threeparttable}

\usepackage{hyperref}
\usepackage{xurl}

\journal{Physica A}

\begin{document}

\begin{frontmatter}



\title{Applying Deep Reinforcement Learning to the HP Model for Protein Structure Prediction}


\author[inst1]{Kaiyuan Yang\fnref{label_ky}}
\fntext[label_ky]{Present address: Department of Quantitative Biomedicine at the University of Zurich, Switzerland}

\affiliation[inst1]{organization={Department of Computer Science, School of Computing, National University of Singapore},
            postcode={117417}, 
            country={Singapore}}

\author[inst2]{Houjing Huang}

\affiliation[inst2]{organization={Institute of Automation, Chinese Academy of Sciences},
            city={Beijing},
            postcode={100190}, 
            country={China}}

\author[inst3]{Olafs Vandans}

\affiliation[inst3]{organization={EXN SIA},
            city={Jurmala},
            country={Latvia}}

\author[inst4]{Adithya Murali}

\affiliation[inst4]{organization={NVIDIA Seattle Robotics Lab},
            city={Redmond},
            postcode={98052}, 
            state={WA},
            country={United States}}

\author[inst5]{Fujia Tian}
\affiliation[inst5]{organization={Department of Physics, City University of Hong Kong},
            addressline={83 Tat Chee Avenue, Kowloon}, 
            city={Hong Kong},
            country={China}}

\author[inst1]{Roland H.C. Yap\corref{cor1}}
\ead{ryap@comp.nus.edu.sg}

\author[inst5]{Liang Dai\corref{cor1}}
\ead{liangdai@cityu.edu.hk}

\cortext[cor1]{Co-Corresponding authors}

\begin{abstract}
A central problem in computational biophysics is protein structure prediction,
i.e., finding the optimal folding of a given amino acid sequence.
This problem has been studied in a classical abstract model, the HP model,
where the protein is modeled as a sequence of H (hydrophobic) and P (polar) amino acids on a lattice.
The objective is to find conformations maximizing H-H contacts.
It is known that even in this reduced setting, the problem is intractable (NP-hard).
In this work, we apply deep reinforcement learning (DRL) to the two-dimensional HP model.
We can obtain the conformations of best known energies for benchmark HP sequences with lengths from 20 to 50.
Our DRL is based on a deep Q-network (DQN).
We find that a DQN based on long short-term memory (LSTM) architecture greatly enhances the RL learning ability and significantly improves the search process.
DRL can sample the state space efficiently,
without the need of manual heuristics.
Experimentally we show that it can
find multiple distinct best-known solutions per trial.
This study demonstrates the effectiveness of deep reinforcement learning in the HP model for protein folding.
\end{abstract}




\end{frontmatter}


\section{Introduction}

\begin{figure}
\centering
    \includegraphics[width=0.5\textwidth]{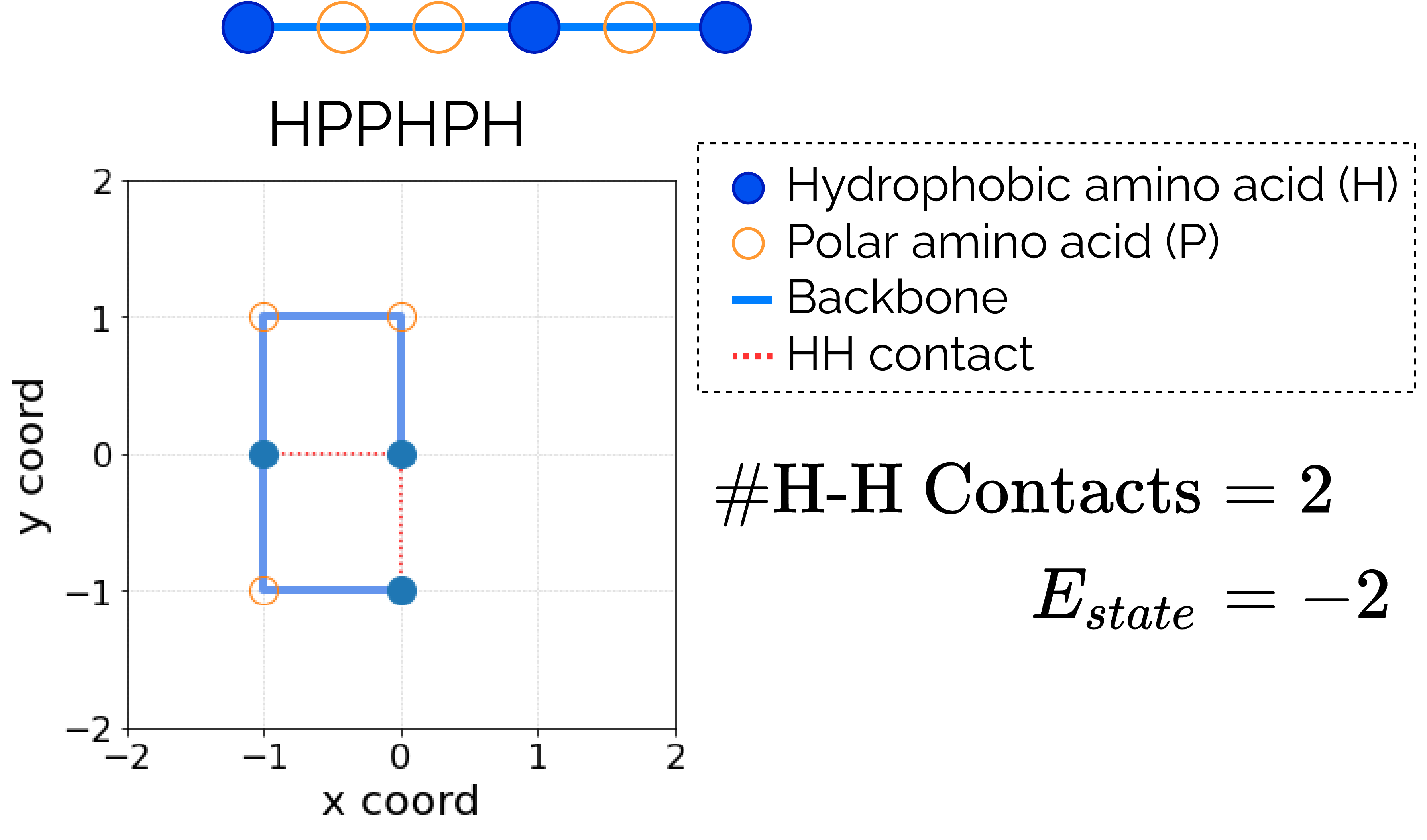}
    \caption{\label{fig:HP_eg_6mer}
        HP model on a 2D square lattice.
        Example six-unit HP sequence,
        $S$=HPPHPH
        with
        $N$=6,
        folds into a conformation with two H-H contacts. The energy of this state is $E_{\text{state}}=-2$ (the optimal conformation).
    }
\end{figure}

Predicting protein structure from a sequence of amino acids is one of the central problems in computational biophysics research~\cite{istrail2009combinatorial, hart2006proteinMIT}.
From the viewpoint of statistical physics, proteins usually fold into the optimal structures with minimum free energies~\cite{anfinsen1973principles}.
However, finding the optimal structure for realistic proteins is too complicated.
Following similar approaches to address other complicated problems, physicists have developed an abstract model, the HP model,
to simplify the protein structure prediction problem~\cite{ken_dill_1985theory}.
In the HP model, a protein is represented as a chain of monomers on a 2D or 3D lattice.
Each monomer can be either H, standing for hydrophobic, or P, standing for polar.
The task is to find the optimal structure for a given sequence of H and P, such as HPPHPH in Fig.~\ref{fig:HP_eg_6mer}.
The optimal structure is defined as the structure with the maximum number of H-H contacts.
The physical reasoning behind the HP model is that protein structural stability is contributed by the attraction between hydrophobic residues to a large extent~\cite{globular_dill1995principles, Review_SWill2000, PHYSICA_A_Tang2000}.

Even though the HP model is already a simplified model for protein folding, it is still hard to find the optimal structure.
Actually, finding the optimal structure in the HP model is NP-hard~\cite{NP_crescenzi1998complexity, demaine_aichholzer2003long, istrail2009combinatorial}.
We highlight that the HP model represents the \textit{ab initio} paradigm heavily rooted in biophysics,
which is different from the AlphaFold~\cite{alphafold2proteins, alphafoldnature2021} and the academia counterpart
`RoseTTaFold'~\cite{RoseTTaFold_baek2021accurate} methods that rely heavily on protein structure data.
The value of \textit{ab initio} approaches is that they may help to better understand the problem.
Many methods have been employed to address the HP model, such as Monte Carlo simulations~\cite{1990MC_Science, EMC2001, HOOS_thachuk2007replica, 2007MC_zhangjingfeng, 2009MC_Landau, PERM1997_grassberger, PERM1998_testing, PERM2003, PHYSICA_A_Landau2021}.
In recent years, the advances of machine learning have helped tackle many classical problems in biophysics with considerable success~\cite{Review_PRE2017Chen, Review_PRL2019Chen, Review_ModePhy2019Carleo, PRE_knot_ovky, Review_JCP2020Zhang, Review_StrucBio2020Noe, Review_PNAS2021JiangYing}.
In particular, reinforcement learning (RL), a paradigm of machine learning~\cite{RLsuttonBook_sutton2018reinforcement}, has been demonstrated to be effective in solving many difficult problems~\cite{mnih2015human, openAIFIVE_berner2019dota, efficientZero_2021, Nature_2022_DRL_nuclearPlant}.
Here, we are motivated to apply RL to the HP model and evaluate its performance.


The task of adopting RL to the HP model is not trivial and comes with a variety of design choices and implementation details.
There are many components and considerations to be made,
and we list a couple of studies that have tested various approaches.
In 2011, Czibula et al.~\cite{SAW_notation_czibula2011reinforcement} employed traditional tabular Q-learning for a very simple HP sequence of HPPH, which was one of the first studies of applying RL to the HP model.
In 2015, Dogan et al.~\cite{AntQ2015} 
modified tabular Q-learning with ant-swarm-based heuristics for longer HP sequences, up to length 48,
and achieved close to best-known solutions for selected benchmark HP sequences.
In more recent studies, deep reinforcement learning (DRL) was applied to the HP model. One of the first works was by Li et al.~\cite{FoldingZero_li2018foldingzero} in 2018, which adopted the architecture from the Go-game-playing AlphaGo Zero~\cite{AlphaGoZero_2017} (thus named ``FoldingZero").
They employed Monte Carlo Tree Search, imitation learning, and a NN with multi-heads in their DRL design.
In 2019, Wu et al.~\cite{Suzhou_BMC_wu2019research}
used tabular RL in the form of classical Q-learning.
Traditional tabular RL is limited by the memory storage size for practical problems.
The states in tabular Q-learning are usually static indices and do not have much generalization capacity~\cite{RLsuttonBook_sutton2018reinforcement}.
Limited by the exponential growth of state-space-size,
Wu et al. carried out experiments on short HP sequences, most of which are shorter than length 20. They
achieved best-known solutions for benchmark sequences with the length of 20.
In 2020, Yu et al.~\cite{2020NIPS_yudeep} applied various DRL methods to the HP model, including deep Q-network (DQN).
They implemented DRL algorithms such as policy iteration, value iteration with dynamic programming as well as various advanced versions of DQN such as prioritized DQN and dueling DQN.
They also tried with AlphaGo Zero with pre-training as another class of DRL methods.

\begin{figure}
\centering
\includegraphics[width=0.65\textwidth]{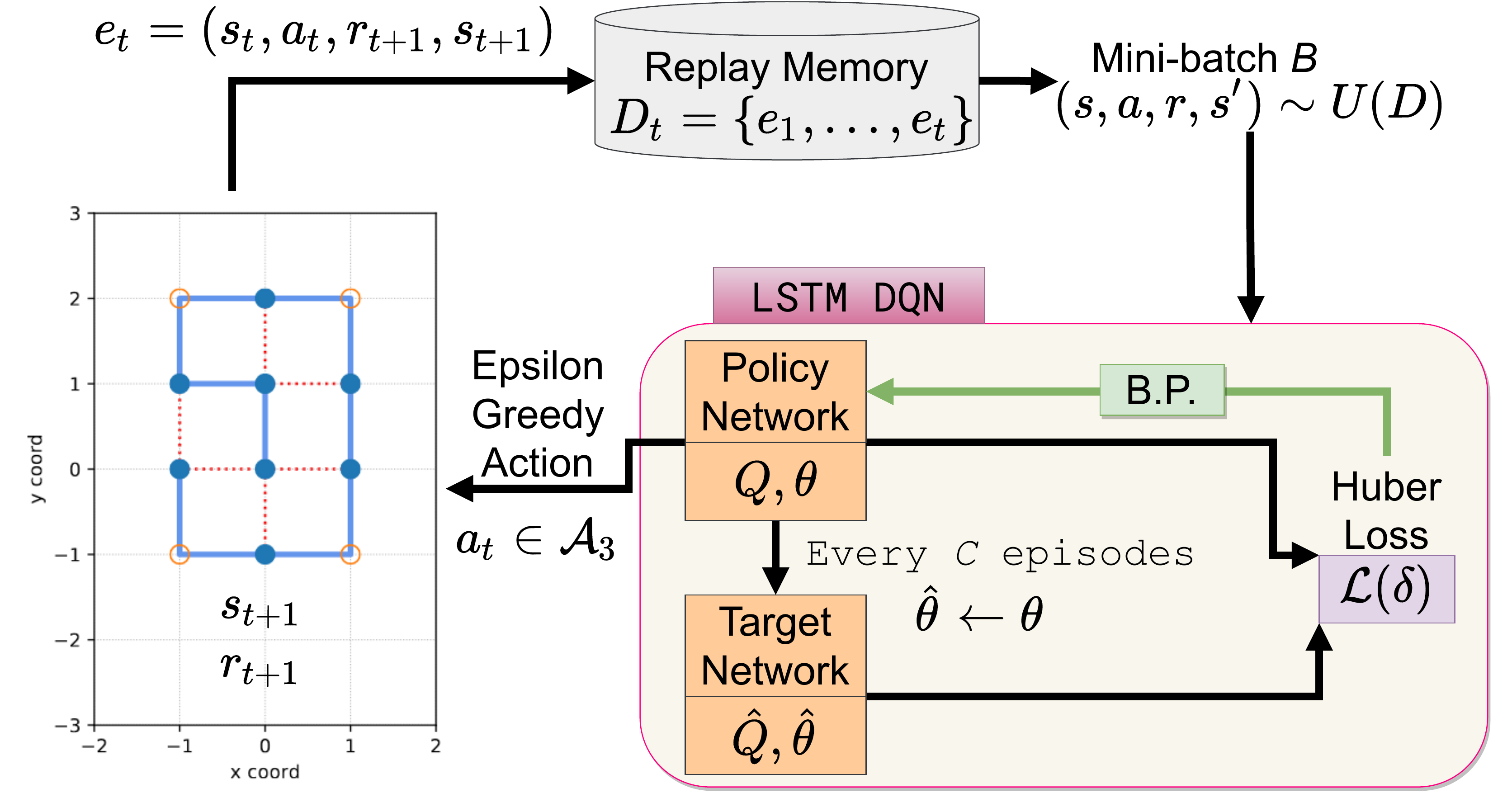}
\caption{\label{fig:EnvOverview}
    Overview of our DRL method with LSTM-based DQN.
    Enclosed in the purple border is the deep NN component.
    There are two NNs for the learning, the Policy Network and the Target Network.
    The TD error is used to calculate the loss.
    Back propagation (B.P. in green) then tunes the learnable parameters $\theta$ in the Policy Network.
    The HP model SAW agent interacts with the RL environment at each time step $t$ by taking actions based on the $\epsilon$-greedy method.
    Action-state-reward experiences $e_t$ are stored in replay memory $D_t$.
    Mini-batches $B$ of experiences are uniformly sampled for DQN training.
}
\end{figure}

In this work, we investigate what ``ingredients” are best needed for a simple ``recipe” to solve the HP model using a DRL setup.
We consider the key components in DRL including the state representation,
set of actions,
reward shaping,
neural network (NN) architecture,
exploration-exploitation trade-off,
and Markov Decision Process (MDP) computation.
Here, we will start from a basic prototypical version of the original DQN~\cite{mnih2013playing, mnih2015human},
then we gradually modify/add/subtract components as needed,
and finally construct a DRL setup as shown in
Fig.~\ref{fig:EnvOverview}
(as a quick overview, and we will elaborate more in our Methodology section).
Our DRL setup for the HP model can achieve conformations of best-known energies for test benchmark sequences with lengths up to 50,
outperforming previous RL approaches on the HP model.

\section{Methodology}

\subsection{HP Model on 2D Square Lattice}

\begin{figure*}
    \includegraphics[width=1.0\textwidth]{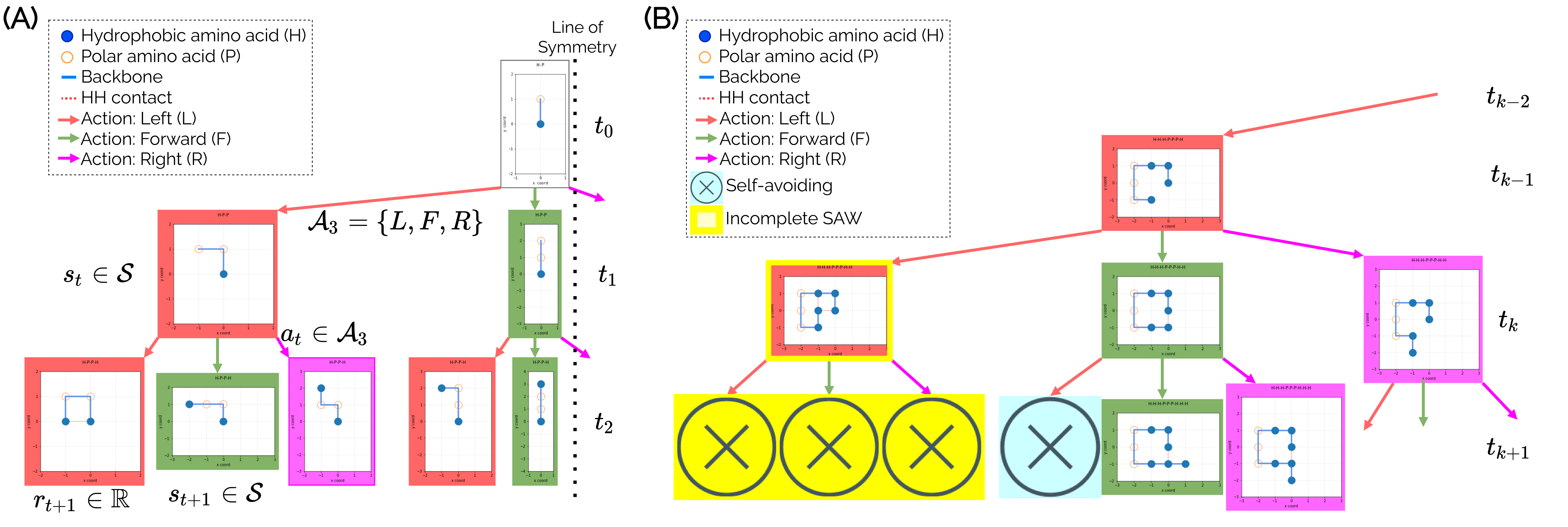}
    \caption{\label{fig:MDP_and_trapped} \textbf{(A)} HP model SAW as a Markov Decision Process.
    At each discrete time step ${t_0, t_1, t_2, ...}$,
    the walker agent takes an action $a_t \in \mathcal{A}_3$,
    places an HP residue on the lattice site,
    and transitions from state $s_t$ to $s_{t+1}$.
    The RL environment interacts with the agent through a (real valued) reward signal $r_{t+1}$ at each time step.
    After the terminal time step $T$, the walker agent completes the SAW,
    a new episode starts
    and the agent resets to the initial default state at $t_0$ all over again.
    \textbf{(B)} Note at time-step $t_{k-1}$, if the SAW agent goes `Left', then the walk is incomplete and terminates at $t_k$!
    An incomplete SAW, highlighted in yellow,
    has no valid actions from $\mathcal{A}_3$, so has no available actions to increase the length.
    Note the self-avoiding constraint, highlighted in cyan,
    forbids the agent from going `Forward' at $t_k$ then `Left' at $t_{k+1}$.
    }
\end{figure*}

The HP model folding process is set up as a self-avoiding walk (SAW).
The HP sequence starts the walk from the origin of the 2D Cartesian plane,
so we place the 1st HP unit (\textit{HP unit is the amino acid monomer in the HP sequence}) on the lattice site with coordinate $(0,0)$.
We ensure our set of SAW is invariant to translational and rotational symmetries by adding the following constraints~\cite{NonSym_bornberg1997model, LUND_irback2002enumerating}.
The first two HP units are put in a position of \textit{`up'}.
And the first turn (non-Forward action) is only allowed to be Left,
see Fig.~\ref{fig:MDP_and_trapped}(A).
Thus the 2nd HP unit is always placed on the coordinate $(0,1)$.
The first two HP units are always fixed with the same starting positions, as shown in the $t_0$ state of Fig.~\ref{fig:MDP_and_trapped}(A).
Each successive HP unit is placed onto the lattice site following this square-grid constraint~\cite{SAW_notation_czibula2011reinforcement}:
\begin{equation}
    \forall i,j \in \{1,...,N\} \ldotp \;
    |i-j| = 1 \implies |x_i - x_j| + |y_i - y_j| = 1,
\end{equation}
where $N$ is the length of the HP sequence $S$, and $(x_i,y_i)$ is the coordinate on the Cartesian plane of the i-th HP unit,
and $(x,y) \in \mathbb{Z} \times \mathbb{Z}$.
Similarly, the self-avoiding constraint can be formulated as~\cite{SAW_notation_czibula2011reinforcement}:
\begin{equation}
    \forall i,j \in \{1,...,N\} \ldotp \;
    i \ne j \iff (x_i, y_i) \ne (x_j, y_j)
\end{equation}

The SAW uses a relative direction scheme consisting of $\mathcal{A}_3 = \{L, F, R\}$.
The SAW is set up as a Growing SAW, or GSAW~\cite{trapped_hemmer1984average, trapped_lyklema1986monte, trapped_hooper2020trapping}.
Basically the SAW grows along the edges of a 2D square lattice by placing the HP units one after the other.
Constrained by the self-avoiding property, not all actions $a \in \mathcal{A}_3$ are valid or available in every time step.
We borrow the term `one-step probability' from~\cite{trapped_lyklema1986monte}: one-step probability is 1 over the number of unoccupied nearest-neighbour sites, and it can take values from \(\{\frac{1}{3}, \frac{1}{2}, 1, 0\}\).
One-step probability of 0 means the SAW is prematurely terminated, or is an incomplete SAW (see Fig.~\ref{fig:MDP_and_trapped}(B)).

A \textit{contact} is formed when two amino acid units are adjacent on the lattice grid sites,
but not adjacent in the HP sequence string, i.e.,not linked via the peptide backbone.
Contacts between two hydrophobic H units are called \textit{H-H contacts}.
The free energy of a conformation (or state), denoted as $E_{\text{state}}$, is the negative value of the number of H-H contacts.
Fig.~\ref{fig:HP_eg_6mer} shows a conformation of an HP sequence ($S$) of HPPHPH with $E_{\text{state}}=-2$.

\subsection{DRL Preliminaries}



\begin{figure*}
    \includegraphics[width=1.0\textwidth]{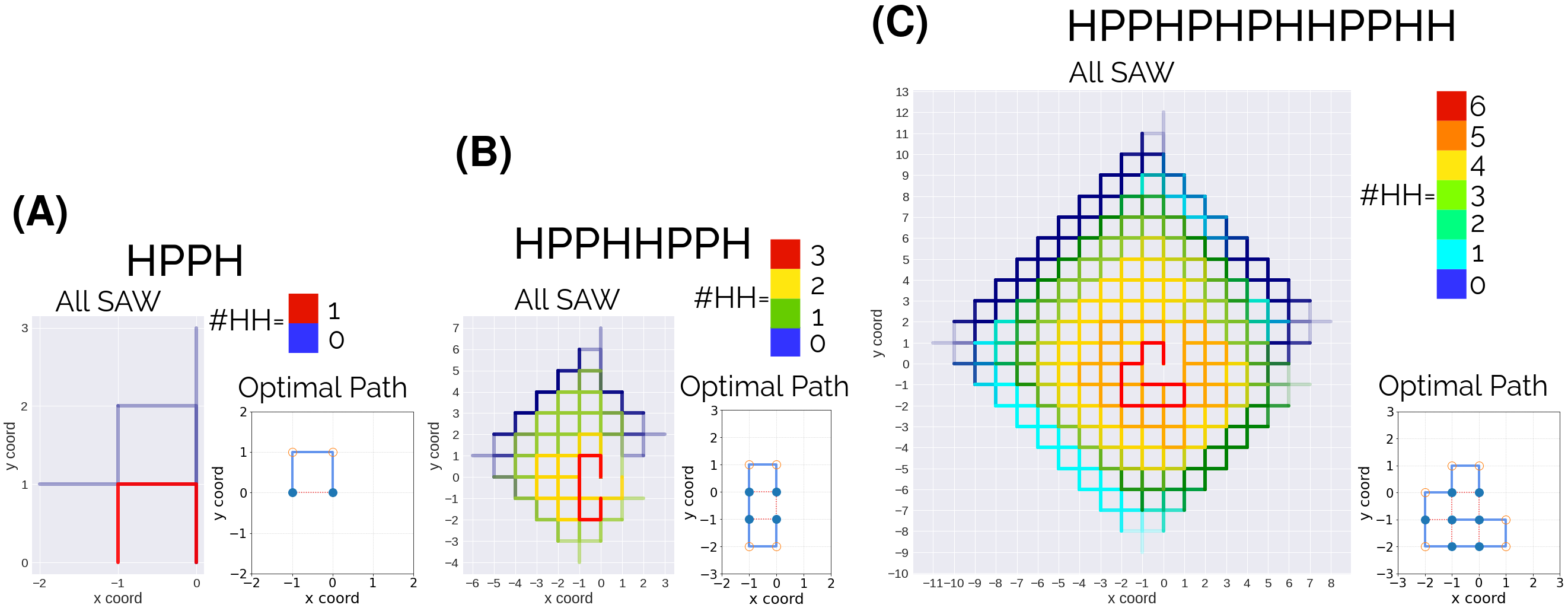}
    \caption{\label{fig:optimal_SAW_heatmap}
        HP model as optimal SAW path finding.
        All possible SAW paths from a HP sequence are overlaid with a red-blue color-scale ``energy heatmap".
        Paths with more number of H-H contacts
        are plotted with less transparency and overlaid on top of paths with lower scores.
        \textbf{(A)} HP sequence of length $N$=4, $S$=HPPH,
        with the red and blue paths having optimal and zero H-H contacts, respectively.
        \textbf{(B)} HP sequence of length $N$=8, $S$=HPPHHPPH.
        The red SAW path has an optimal score of 3, giving
        a  folding with three H-H contacts.
        \textbf{(C)} HP sequence of length $N$=13, $S$=HPPHPHPHHPPHH.
        The red-blue heatmap is red with score of 6 and blue with score of zero.
    }
\end{figure*}

In this study, we view solving the HP model as optimal SAW path finding to optimize the HP model score or number of H-H contacts.
Fig.~\ref{fig:optimal_SAW_heatmap} illustrates our perspective.
For each HP sequence,
the `folding' is a SAW path as the HP backbone and its comprising units are embedded in the lattice grid in a non-overlapping fashion.
In Fig.~\ref{fig:optimal_SAW_heatmap},
we overlay all SAW made by the sequence with a heatmap showing the number of H-H contacts.
The more H-H contacts, the higher the score assigned to a particular SAW path, and the ``hotter" the temperature on the graph.
Note the \textbf{sequential decision making} nature of the SAW---
the optimal HP model folding is a sequence of walk-steps that traces out the optimal SAW path.
Thus, the problem can be framed as a Markov Decision Process (MDP) and solved by approaches from optimal control theory, in particular, reinforcement learning (RL).



RL seeks to maximize a numerical reward signal by figuring out which action to take from a particular state~\cite{RLsuttonBook_sutton2018reinforcement}.
Without explicit and immediate supervision, the agent discovers the optimal way to act in an environment by trying actions on its own.
In our setting,
the RL agent is the HP model SAW walker that tries to walk a path to obtain the maximum reward.
The length of the SAW path is determined by the HP sequence length, $N$.
Our RL environment is the 2D square lattice on a Cartesian plane that sends out rewards.
The reward is associated with the free energy of the folded state as calculated by the number of H-H contacts,
and also other reward shaping to be discussed later.

In our MDP formulation, we adopt a discrete time step setting.
At each time step $t$, the agent and environment interact.
The agent receives an observation of the state, $s_t \in \mathcal{S}$, where $\mathcal{S}$ is the set of all states of the observed environment.
Based on the observed state $s_t$, the agent takes an action at that time step, $a_t \in \mathcal{A}_3$, where $\mathcal{A}_3$ is the set of all three possible actions for a SAW, namely to go left, forward or right.
The environment reacts to the agent and sends out a real number value as reward, $r_{t+1} \in \mathbb{R}$.
The agent also observes a new state of the environment, $s_{t+1}$.
Fig.~\ref{fig:MDP_and_trapped}(A) provides the schematic view of a SAW as MDP.

To optimize the MDP of the HP model SAW,
RL is used to determine the long-term desirability of states with a \textbf{value function}.
Values are essential to the RL agent to make action decisions, as value judgement of states predicts the amount of cumulative reward over time.
Formally, the value function $Q(s, a)$ is the expected return for taking action $a$ while being in state $s$.
The \textit{optimal} value function $Q_*$ is related to the \textit{Bellman optimality equation}~\cite{RLsuttonBook_sutton2018reinforcement}:
\begin{equation}
    Q_*(s, a) = \mathbb{E} \left[ r_{t+1} + \gamma
    \max_{a^\prime} Q_*(s^\prime, a^\prime)
    \,\middle\vert\,  s_t = s, a_t=a \right],
    \label{eq:bellman_optimality_Q_eq}
\end{equation}
where $\gamma$ is the discount factor, $0 \leq \gamma \leq 1$, and $s^\prime=s_{t+1}$, $a^\prime=a_{t+1}$ represent the next state and next action.
The optimal value function is non-linear,
and is typically \textit{approximately solved} with iterative methods~\cite{RLsuttonBook_sutton2018reinforcement},
such as \textbf{Q-learning}.
Q-learning approximates the optimal value function $Q_*$.
The basic procedure for Q-learning is to loop through a number of \textbf{episodes} (\textit{Finite horizon MDP can end in a terminal state, after which the agent resets. Such a finite sequence of decisions is called an episode.}),
and for each step in an episode, the Q-learning agent takes action $a$ and observes $r, s^\prime$,
while updating $Q$ by:
\begin{equation}
    Q(s, a) \gets Q(s, a) +
    \beta \left(
    r + \gamma \max_{a^\prime} Q(s^\prime, a^\prime) -
    Q(s, a) \right), \label{eq:Q-learning-Eq}
\end{equation}
where $\beta$ is the step-size parameter, $0< \beta < 1$.
Q-learning is a type of \textbf{temporal-difference} (TD) method because the learning is based on the differences between two estimates, $Q_{t+1} - Q_t$, at two successive time steps.

For RL tasks with small sets of states and actions, a lookup table for state-action pairs (as in \textit{tabular RL}) can handle the approximation of the optimal value function.
However, the memory storage required for such tabular methods can quickly
become prohibitive.
Artificial neural networks (NN) have been a highly useful and effective non-linear function approximator~\cite{RLsuttonBook_sutton2018reinforcement}.
NNs with multiple hidden layers are called deep NNs and define the practice of \textit{deep learning} (DL)~\cite{goodfellow2016deep}.
RL that has a DL component is called \textbf{deep reinforcement learning} (DRL).
In DRL, the approximation of optimal value function is in the form of a parameterized function approximator, i.e., the NN, as opposed to a table in tabular RL (such as in classical Q-learning).

One of the pioneering works to combine DL with RL and demonstrate DRL's success is Mnih et al~\cite{mnih2013playing, mnih2015human}.
Their algorithm uses a deep NN to estimate the value function, and is called a \textbf{deep Q-network} (DQN).
The objective of the NN in DQN is to better approximate the optimal value function $Q_*(s, a)$.
The Bellman optimality
Eq.~(\ref{eq:bellman_optimality_Q_eq}) for the value function is used as the target value to train the NN.
The difference between the NN output Q-values and the optimal Q-values is the temporal-difference or \textbf{TD error}, $\delta$:
\begin{equation}
    \delta = \left(
        r + \gamma \max_{a^\prime} Q(s^\prime, a^\prime)
    \right)
    -
    Q(s, a)
    \label{eq:TD_error_eq}
\end{equation}
NN attempts to minimize the TD error $\delta$ during training.

The NN architecture and design of hidden layers affect the features learned.
Various NN architectures have been proposed to solve specific machine learning tasks,
in particular,
recurrent neural networks (RNN) for time-series and sequential data~\cite{goodfellow2016deep, PRE_knot_ovky}.
Another refined version of RNN called \textit{long short-term memory} (LSTM)
was proposed by Hochreiter and Schmidhuber~\cite{lstm_hochreiter1997long},
which specifically addresses the limited short-term memory faced by basic RNNs.
These improvements over plain RNN make LSTM a suitable architecture to detect long-term dependencies in sequential and time-series data.
Accordingly, the walking steps in the HP model can be seen as a sequence of actions in time.
We hypothesise that using LSTM as the NN architecture in the DQN can provide a representation that captures long range interactions in the folding and improves the learning.

\subsection{Our DRL Setup}

Fig.~\ref{fig:EnvOverview} gives an overview of our DRL method using DQN.
The agent interacts with the environment by taking actions and acquiring experiences,
which are collected into the replay memory;
then the LSTM NN samples from the replay memory with mini-batches for training
and minimizing the TD error loss to approximate the optimal value function.
We now describe the individual key components in our DRL setup as follows.

\subsubsection{Exploration and Exploitation}

Since RL learns through trial-and-error and the agent `discovers' a good policy from its experience of the environment,
it is important for the agent to both \textit{explore} and \textit{exploit}.
The agent explores the environment to discover new reward from new actions,
and also exploits past experience that was found to be effective.
To balance exploration and exploitation, the agent tries more new actions in the beginning of an episode, and progressively utilizes what it has already experienced.

The $\epsilon$-greedy strategy is a simple scheme to ensure the agent has continual exploration.
All actions in all states are tried with non-zero probability.
$\epsilon$ is the probability that an action $a$ is chosen at random uniformly:
\begin{equation}
  a=\begin{cases}
    \arg\,\max_{a \in \mathcal{A}_3} Q(s,a), & \text{with probability ($1-\epsilon$)};\\
    a \sim U(\mathcal{A}_3), & \text{with probability $\epsilon$}.
  \end{cases}
\end{equation}

In this study, we balance exploration and exploitation using $\epsilon$-greedy approach,
where $\epsilon$ is the probability of selecting an action at random.
We also compare with a strategy that randomly explore the state space throughout, called `RAND'.
In `RAND',  $\epsilon$ is set to be a constant, $\epsilon = \epsilon_{max} = 100\%$ for all episodes.
For RL based experiments, $\epsilon$ decays following an exponential schedule,
according to the following formula:
\begin{equation}
    \epsilon_i = \epsilon_{min} + \left( \epsilon_{max} - \epsilon_{min} \right)
    \exp\left({\frac{-i\lambda}{\psi}}\right),
    \label{epsilon_decay_eq}
\end{equation}
where $\epsilon_i$ is the exploration rate for episode with index $i$, and $\psi$ is the total number of episodes for decay,
$\lambda=5$ is the decay rate,
$\epsilon_{min}=0.01$ and $\epsilon_{max}=1$ are the minimum and maximum values set for the exploration rate.

\subsubsection{Reward Function}

Our RL environment has a sparse-reward setting.
The agent receives non-zero reward only at the end of a finished SAW episode, i.e., after the terminal time step $T$ for an episodic walk.
All SAW episodes can finish in one of two ways for the RL agent: complete SAW or incomplete SAW:
\begin{itemize}
    \item \textit{Complete SAW:} SAW that completes the walk by placing all the $N$ HP units onto the lattice.
    \item \textit{Incomplete SAW:} 
    a prematurely terminated SAW shorter than length $N$ because no valid actions are available to maintain the SAW property (see Fig.~\ref{fig:MDP_and_trapped}(B)).
\end{itemize}
Note that \textbf{an incomplete SAW is not a valid solution} (see Fig.~\ref{fig:MDP_and_trapped}(B)).

For an HP sequence $S$,
with each state-action transition to a new state $s_{t+1}$,
the RL environment computes the number of H-H contacts of $s_{t+1}$ or $\lvert E_{\text{state}} \rvert$.
The reward function is as follows:
\begin{equation}
    r_{t+1} = \eta \cdot \lvert E_{\text{state}} \rvert \label{reward_func}
\end{equation}
\begin{equation*}
  \eta=\begin{cases}
    1, & \text{if episode is finished, i.e. } t=T.\\
    0, & \text{otherwise}.
  \end{cases}
\end{equation*}

\subsubsection{State Representation}

\begin{figure}
\centering
\includegraphics[width=0.5\textwidth]{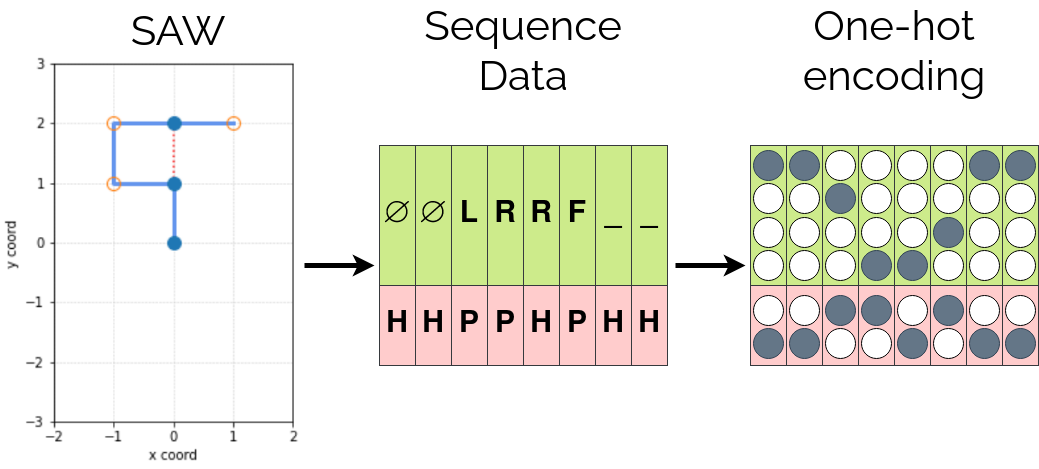}
\caption{
The SAW on the left is not yet complete as there are still two more HP units to be placed onto the 2D square lattice.
The current SAW state is converted into a sequence vector of length $N$ representing movement status and monomer type.
The first two HP units are fixed, so the action associated is `$\varnothing$'.
The remaining two HP units are not yet placed on the lattice and are indicated by place-holder `\_' to mean actions to be determined.
The action sequence for the current SAW state is $L, R, R, F$.
Finally we encode the sequence
data with one-hot encoding for 6 features to preserve their categorical properties.
}
\label{fig:one_hot_state}
\end{figure}

The input to the NN in DQN is a one-hot encoded vector~\cite{geron2019hands} representing the state,
as shown in Fig.~\ref{fig:one_hot_state}, which gives an example for an 8mer HHPPHPHH.
One-hot encoding is applied to represent both the action sequence as well as the HP sequence,
which results in a six-dimensional binary vector.
The six-dimensional features are made up of the four movement status types (-,L,F,R) and two monomer types (H,P).
The vector length is fixed to be $N$ matching the $N$-mer,
and HP units yet to be placed on the lattice have movement code `-'.
These one-hot encoded arrays are then fed into the RNN of the DQN.

\subsubsection{Neural Network Architecture}

\begin{figure}
\centering
\includegraphics[width=0.5\textwidth]{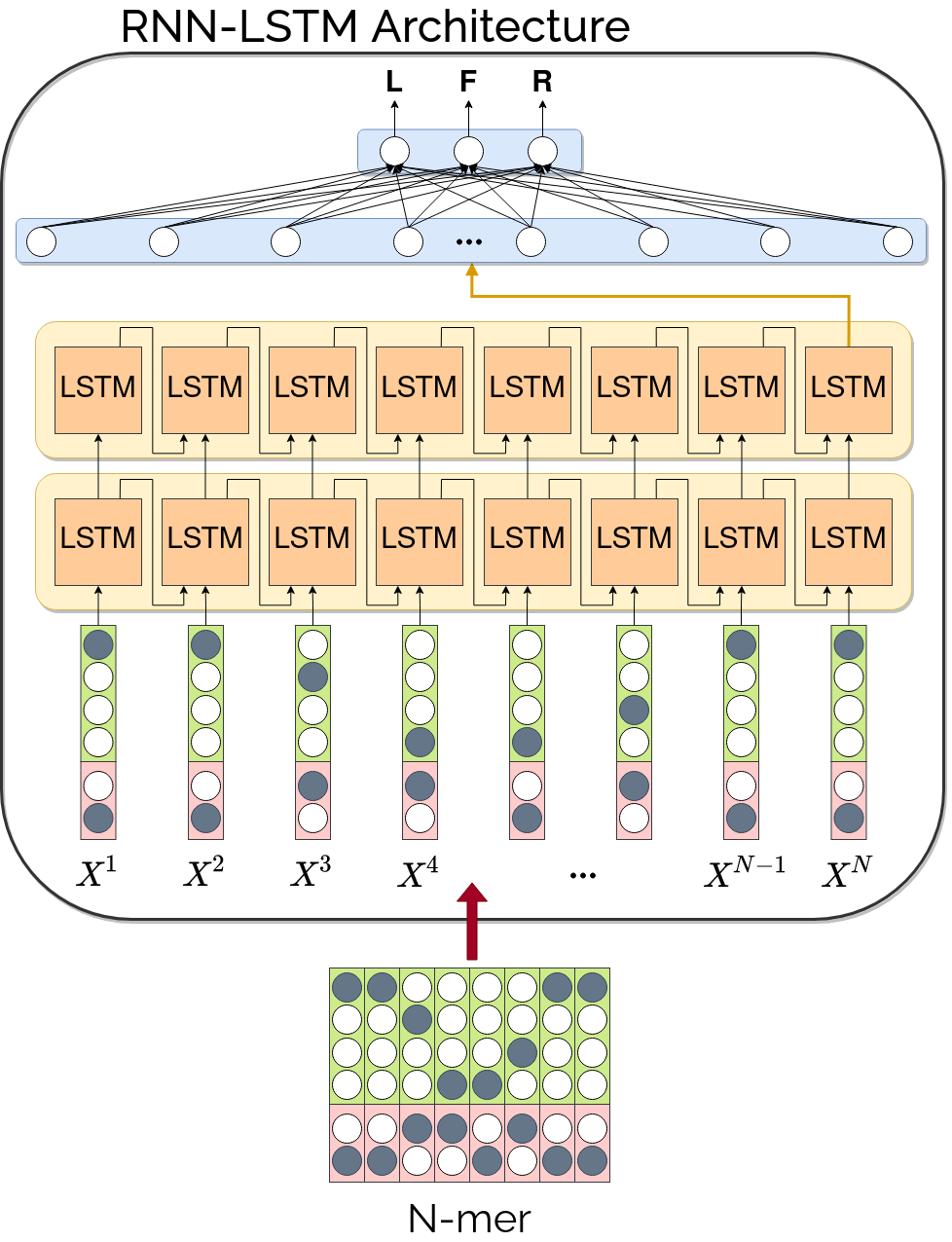}
\caption{LSTM-based recurrent neural network used in DQN. Input $X^{(t)}$ represents the t-th step of input sequence $X$ with length $N$. The first layer of LSTM cells accepts 6-dimensional inputs, and produces outputs which are passed downstream. Data flows from stacked LSTM layer to layer, as well as between cells within layers, capturing sequentiality.
Hidden state of the final recurrent layer's N-th cell is passed to another fully connected layer.
The last output layer outputs the predicted Q-values for the three actions $a \in \mathcal{A}_3$.}
\label{fig:lstm_arch}
\end{figure}

The LSTM neural network takes in a one-hot encoded vector while preserving the sequential nature of the state representation.
Fig.~\ref{fig:lstm_arch} illustrates the architecture of our LSTM NN.
For a $N$-mer with $N$ columns of one-hot encoded vectors $X^1, ..., X^N$,
each $X$ is an input with 6 features,
which are 4 movement (-,L,F,R) features plus the 2 residue (H,P) features.
Sequentially $X^1, ..., X^N$ is fed into the first LSTM layer.
The second stacked LSTM layer takes in the outputs of the first LSTM layer and computes the output of learned features in the hidden state.
The last output layer is fully connected to the feature output from the last hidden state of the stacked LSTM layers, and generates Q-values for the three actions.
Note that the LSTM network outputs the Q-values based on the sequential $N \times 6$ input, which captures both the actions performed so far as well as the sequence information of the whole chain as illustrated in Fig.~\ref{fig:one_hot_state}

Through empirical testing,
we found a two-layer 256-hidden-state stacked LSTM architecture is sufficient for HP sequences shorter than $N\leq36$.
For longer HP sequences ($N=48, 50$),
a three-layer LSTM with 512 hidden states can achieve better results.

\subsubsection{Policy Network \& Target Network}

Following the original DQN DeepMind paper~\cite{mnih2015human},
we use two networks to stabilize the learning.
The first network is called the \textit{policy network}.
The policy network is online and updated at every iteration using back propagation.
The agent takes action based on the Q-value and predictions from the policy network.
The second network is identical to the policy network,
but is only updated periodically:
the \textit{target network}.
Outputs from the target network are used as learning targets for the loss function.
Weights of the target network is updated by cloning the policy network's weights over.
Let $Q$ be the policy network,
and $\hat{Q}$ be the target network.
After every $C$ updates to $Q$,
we clone policy network $Q$'s current weights and biases $\theta$ into the target network $\hat{Q}$'s parameters $\hat{\theta}$.
Having a duplicated network has been shown to make the Q-value training more stable~\cite{mnih2015human}.

We rewrite the TD error from
Eq.~(\ref{eq:TD_error_eq}) to reflect the two networks, policy network $Q$ and target network $\hat{Q}$ in the DQN:
\begin{equation}
    \delta = \left(
        r + \gamma \max_{a^\prime} \hat{Q}(s^\prime, a^\prime; \hat{\theta})\right)
        -
        Q(s, a; \theta),
    \label{eq:TD_error_new}
\end{equation}
where $\theta$ and $\hat{\theta}$ are the parameters of the policy and target networks respectively.
We set $C$ to be 100 in our DRL experiments.
The discount factor $\gamma$ is set to be $0.98$.

\subsubsection{Replay Memory}

The technique of \textit{replay memory} dates back to Lin in 1992~\cite{RLsuttonBook_sutton2018reinforcement, experiencereplay_lin1992self}.
The original DeepMind DQN paper~\cite{mnih2015human}, however, is one of the first to use experience replay memory in a DRL setting.
Replay memory is simply a storage of past experiences of the agent.
We denote the agent's experience at time step $t$ as a tuple $e_t = (s_t, a_t, r_{t+1}, s_{t+1})$.
The replay memory at time step $t$ is a set containing the stored experiences, $D_t = \{e_1, ..., e_t\}$.
We store the last $\min(50000, \psi/10)$ number of time steps of transitions in the replay memory buffer structured as a first-in-first-out queue and
$\psi$ is the total number of episodes per trial.

\subsubsection{TD Error \& Huber Loss}

The objective of the NN in DQN is to minimize the parameterized TD error $\delta$ in
Eq.~(\ref{eq:TD_error_new}) during training with \textit{mini-batch gradient descent}.
The loss and gradient calculations are done over a small batch of samples.
The mini-batch is denoted as $B$.
The loss function associated with mini-batch gradient descent training is:
\begin{equation}
    \text{loss} = \frac{1}{|B|} \sum_{e_i \in B} \mathcal{L}(\delta),
\end{equation}
where $\mathcal{L}$ is Huber loss~\cite{geron2019hands}, which is a variant of mean square error (MSE) loss function, defined as:
\begin{equation}
  \mathcal{L}(\delta) = \begin{cases}
    \frac{1}{2} \delta^2, & \text{if $|\delta| \leq 1$},\\
    |\delta| - 0.5 & \text{otherwise}.
  \end{cases}
\end{equation}

The mini-batches for training are sampled uniformly at random from the replay memory, $(s, a, r, s^\prime) \sim U(D)$.
Our mini-batch size used in DRL training is 32.
The optimizer used for learning is Adam~\cite{lr_kingma2017adam},
with a learning rate of $0.0005$.

\subsection{Benchmark HP Sequences}

Conveniently, Istrail's research group from the Brown University has been maintaining a popular benchmark sequence set documenting the best-known lowest energy folding for various HP sequences.
These benchmark sequences have been referred to by numerous publications in the field \cite{wuhanBB_chen2005branch, HOOS_thachuk2007replica}, and are included in all the related work mentioned previously.

For the DRL experiments, $N$-mers, where $N \in \{20, 24, 25, 36,48,50\}$, are selected from the Istrail benchmark for comparison with related work.
Table~\ref{tab:table_istrail_benchmark_info} shows the best-known energies of selected lengths and their exact sequences.
\begin{table*}
\caption{\label{tab:table_istrail_benchmark_info}%
    Selected benchmark HP sequences from the Istrail Benchmark with their best known energies.
}
        \centering
        \resizebox{1.0\linewidth}{!}{
\begin{tabular}{lclc}
\hline
\textrm{$N$-mer ID}&
\textrm{Length}&
\textrm{HP Sequence}&
\textrm{Best Known Energy}\\
\hline
20mer-A & $20$ & HPHPPHHPHPPHPHHPPHPH & $-9$ \\
20mer-B & $20$ & HHHPPHPHPHPPHPHPHPPH & $-10$ \\
24mer & $24$ & HHPPHPPHPPHPPHPPHPPHPPHH & $-9$ \\
25mer & $25$ & PPHPPHHPPPPHHPPPPHHPPPPHH & $-8$ \\
36mer & $36$ & PPPHHPPHHPPPPPHHHHHHHPPHHPPPPHHPPHPP & $-14$ \\
48mer & $48$ & PPHPPHHPPHHPPPPPHHHHHHHHHHPPPPPPHHPPHHPPHPPHHHHH & $-23$ \\
50mer & $50$ & HHPHPHPHPHHHHPHPPPHPPPHPPPPHPPPHPPPHPHHHHPHPHPHPHH & $-21$ \\
\hline
\end{tabular}
}
\end{table*}

\subsection{Implementation}


The experiments on DRL are run on servers with NVIDIA GeForce RTX 3070 Ti GPUs.
NNs are implemented with the PyTorch framework version 1.10.1, and run on CUDA version 11.3.
Please refer to the setup and installation guide in our GitHub repository (link in Data Availability Statement).
We use the open source Python library OpenAI Gym to develop the RL environment~\cite{openAI}.
Our HP model on a 2D square lattice RL environment is extended and refactored based on the open source repository `gym-lattice'~\cite{miranda2018gymlattice}.
All experiments are repeated four times with four random seeds to ensure reproducibility.

\section{Results \& Discussions}

\subsection{Search Process with DRL}

\begin{figure*}
\centering
\includegraphics[width=0.96\textwidth]{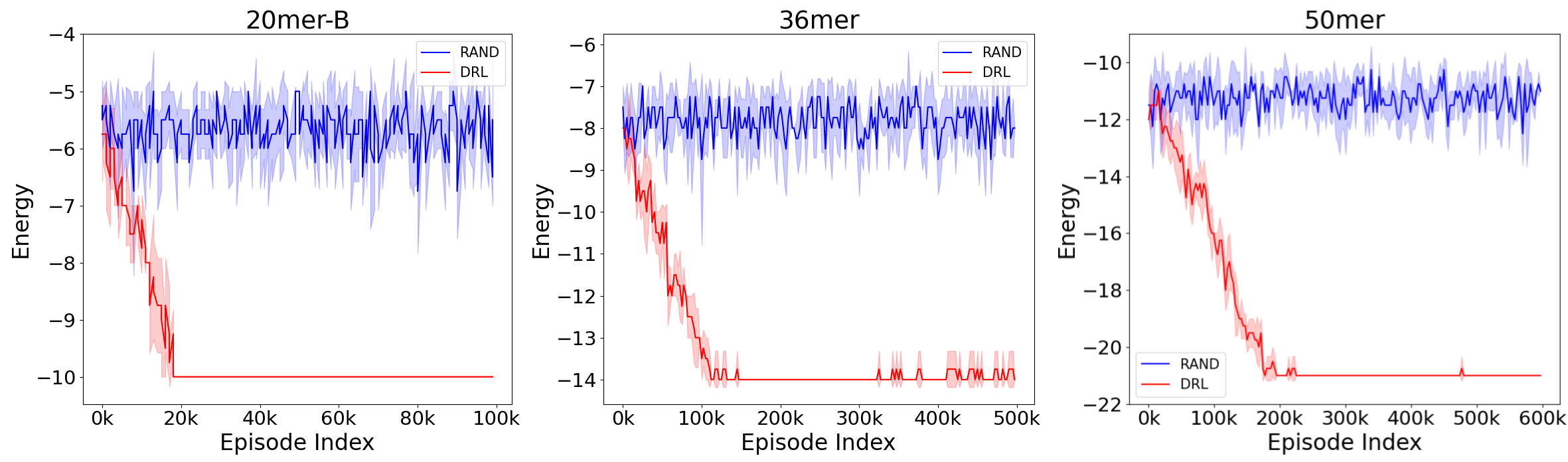}
\caption{\label{fig:epi_curve}
    Learning curves of DRL (red) and RAND (blue) on the 100K-episode-trial of 20mer-B, 500K-episode-trial of 36mer, and 600K-episode-trial of 50mer.
    X-axis is the index of episodes in progression.
    Y-axis denotes the energy found by the agent.
    The curves are plotted with the moving minimum of 200 episodes.
    The shaded area represents the standard deviation.
    Note the best-known minimal energies for 20mer-B, 36mer, and 50mer are $-10$, $-14$, $-21$ respectively.
    Experiments are repeated four times on four random seeds for both RAND and DRL methods.
}
\end{figure*}

For each of the selected benchmark HP sequences from Table~\ref{tab:table_istrail_benchmark_info},
we first establish the baseline performance with pure random explorations, called `RAND' experiments.
For the `RAND' experiments, the HP sequence $N$-mer is allowed to randomly explore, pick valid actions at random during each time step, and grow the SAW.
In contrast to `RAND',
in our DRL experiments, the $N$-mer chooses actions based on the $\epsilon$-greedy algorithm and balances exploration with exploitation.
Fig.~\ref{fig:epi_curve} shows the search process in terms of learning curves of RAND and DRL.

For the shortest $N$=20 of the benchmark sequences,
we run 100K episodes per trial.
Fig.~\ref{fig:epi_curve} displays the search process of one of the 20mers, 20mer-B, as an example.
RAND can discover some close to best-known solutions largely by chance, whereas DRL displays an effective search process,
as shown in the downward curve in red in Fig.~\ref{fig:epi_curve},
and consistently finds the best-known solution of $-10$ in the exploitation phase.

For 36mer, for 500K episodes per trial,
DRL can find the best-known energy of $-14$ while RAND by luck is able to reach $-12$.
Again the moving minimum of the total episodes (red curve in Fig.~\ref{fig:epi_curve}) for DRL shows a clear trend towards a good solution.

For 50mer, we use 600K episodes per trial,
and RAND can only encounter energy of $-15$,
while DRL is able to find the best-known energy of $-21$ in all of the four random seed trials.

Table~\ref{tab:table_perf_comparison} shows the performance of the search process on the benchmark HP sequences from Table~\ref{tab:table_istrail_benchmark_info}.
\begin{table*}
\caption{\label{tab:table_perf_comparison}%
    Performance comparison among related RL work on Istrail Benchmark sequences.
    Table entries are lowest energies of $E_{\text{state}}$ obtained.
    `-' indicates information not provided.
    Results matching the best known energies are highlighted in red bold.
    Second-closest results are highlighted in blue.
}
\begin{threeparttable}
        \centering
        \resizebox{1.0\linewidth}{!}{
\begin{tabular}{lccccccc}
\hline
\textrm{$N$-mer ID}&
\textrm{Dogan-AntQ (2015)}&
\textrm{Li-FoldingZero (2018)}&
\textrm{Wu-QL (2019)}&
\textrm{Yu-DRL (2020)\tnote{a}}&
\textrm{Random}&
\textrm{\textbf{Ours}}&
\textrm{Best Known}\\
\hline
20mer-A & - & \textcolor{red}{$\mathbf{-9}$} & \textcolor{red}{$\mathbf{-9}$} & $-6$ $\vert$ $-8$ & \textcolor{red}{$\mathbf{-9}$} & \textcolor{red}{$\mathbf{-9}$} & $-9$ \\
20mer-B & - & - & \textcolor{red}{$\mathbf{-10}$} & $-8$ $\vert$ \textcolor{blue}{$-9$} & \textcolor{blue}{$-9$} & \textcolor{red}{$\mathbf{-10}$} & $-10$ \\
24mer   & \textcolor{red}{$\mathbf{-9}$} & \textcolor{blue}{$-8$} & - & $-6$ $\vert$ \textcolor{blue}{$-8$} & \textcolor{red}{$\mathbf{-9}$} & \textcolor{red}{$\mathbf{-9}$} & $-9$ \\
25mer   & - & \textcolor{blue}{$-7$} & - & - $\vert$ \textcolor{blue}{$-7$} & \textcolor{blue}{$-7$} & \textcolor{red}{$\mathbf{-8}$} & $-8$ \\
36mer   & \textcolor{blue}{$-13$} & \textcolor{blue}{$-13$} & - & - $\vert$ \textcolor{blue}{$-13$} & $-12$ & \textcolor{red}{$\mathbf{-14}$} & $-14$ \\
48mer   & \textcolor{blue}{$-19$} & $-18$ & - & - & $-17$ & \textcolor{red}{$\mathbf{-23}$} & $-23$ \\
50mer   & - & \textcolor{blue}{$-18$} & - & - & $-15$ & \textcolor{red}{$\mathbf{-21}$} & $-21$ \\
\hline
\end{tabular}
}
        \begin{tablenotes}
        \footnotesize
            \item[a] Yu et al.\cite{2020NIPS_yudeep} reported two classes of RL methods, DRL and AlphaGo Zero with Pretraining. Here the two results \\are separated by `$\vert$'.
        \end{tablenotes}
     \end{threeparttable}
\end{table*}
The number of episodes per trial for $N$-mer is: 100K for 20mer A and B, 500K for 24mer to 36mer, and 600K for 48mer and 50mer.
In Table~\ref{tab:table_perf_comparison} right-most columns,
we list the minimum energy found by RAND, DRL, and best-known results for benchmark HP sequences.
Results matching the best-known energy are highlighted in bold red.

RAND alone can match the best-known results for 20mer-A and 24mer,
and gives close to best-known results for 20mer-B and 25mer.
This shows that RAND can be effective for short sequences with reasonable number of episodes per trial.
The performance gap between DRL and RAND is seen from $N\geq36$.
Our DRL method can find the best-known solutions in all the selected benchmark sequences,
which RAND does not.

\subsection{Performance Comparison with Related RL Work}

As mentioned in related work,
there are a number of works using RL and DRL for the HP model.
We compare our DRL performance with their reported results in Table~\ref{tab:table_perf_comparison}.

Dogan-AntQ (2015)~\cite{AntQ2015}
combines tabular RL with ant-swarm-based heuristics.
Their action set is an absolute direction scheme (left, right, up, down), whereas we use a more compact relative three-direction scheme.
Their reported RL results for 24mer, 36mer and 48mer are $-9$, $-13$, and $-19$ respectively,
which are all short of the best-known results except for 24mer.
For the 36mer and 48mer, they have the second-best performance compared with other related work.

Li-FoldingZero (2018)~\cite{FoldingZero_li2018foldingzero} is one of the first DRL studies on the HP model.
They use a three-action scheme similar to ours.
Li et al. mirror their architecture from the Go-game-playing AlphaGo Zero~\cite{AlphaGoZero_2017} (thus the name FoldingZero).
They are unable to reach the best-known solutions for the benchmark sequences except for the 20mer-A.
We highlight that `RAND' can already encounter best-known solutions for the 24mer in our 500K-episode trials.
Their  result for 24mer is $-8$ and is in the range of `RAND'.
The results for longer sequences are not as good as AntQ and our simpler DQN method.

Wu-QL (2019)~\cite{Suzhou_BMC_wu2019research} is a study using tabular RL.
They use 5 million episodes per trial for all their RL experiments, which may be excessive since it is more than the total number of possible conformations of their short HP sequences (most are shorter than $N$=20).
Using 5 million episodes,
they are able to reach best-known solutions for 20mer-A and 20mer-B.
Due to the exponential growth of number of states, it is difficult for 
tabular RL like Q-learning to deal with longer sequences.
This probably explains why the Wu-QL stops at around $N$=20.

Yu-DRL (2020)~\cite{2020NIPS_yudeep} uses various popular DRL algorithms, including basic DQN, prioritized DQN and dueling DQN.
They also try with AlphaGo Zero with pretraining.
In Table~\ref{tab:table_perf_comparison} their reported results from both DRL schemes are listed.
Their basic DRL methods are in the range of `RAND',
while their AlphaGo Zero with pretraining results are within the range of `RAND' for shorter sequences, and do not reach best-known solutions for the benchmark sequences.

We remark that the related work has not focused on whether their foldings are complete or incomplete SAWs.
Our results are all complete SAW.

Our DRL design follows the original DQN structure from DeepMind paper~\cite{mnih2015human},
and we show that the prototypical DQN algorithm is sufficient to be applied on optimal SAW path finding tasks.
But different from previous work's attempt using DQN (and advanced variants of DQN),
we design a suitable state representation and NN architecture to better utilize the inherent capability of DQN.
Our LSTM-based DQN is able to reach best-known solutions for
all selected benchmark sequences,
and achieves better performance on the HP model than previous results in the literature.

\subsection{The Role of LSTM}

DQN has arguably become the `grandparent' of DRL algorithms since its original proposal in 2013~\cite{mnih2013playing}.
The original DQN papers by Mnih et al.~\cite{mnih2013playing, mnih2015human} showed the first major success of combining NNs and RL in DRL for practical-scale problems,
and launched the field of DRL.
Although since 2013, many add-ons and patches based on DQN have been suggested,
the basic version of DQN is still powerful despite its simplicity.

We think that one of the key reasons why our DRL is successful is the LSTM NN architecture of our DQN.
Li-FoldingZero (2018) and Yu-DRL (2020)  both also use DRL.
In Li et al. their state representation is an image of the lattice board,
and the NN architecture is a convolutional NN (CNN) to learn the local spatial features of the grid.
Yu et al. have tried representing the lattice state both as an image or as a flattened vector.
Subsequently they try to use CNN and fully-connected networks on image and flattened vector representations respectively.
Yu et al. report no significant differences between the two state representations and NN architectures.
Both works have not considered a RNN based NN architecture.

\begin{figure}
\centering
\includegraphics[width=85mm]{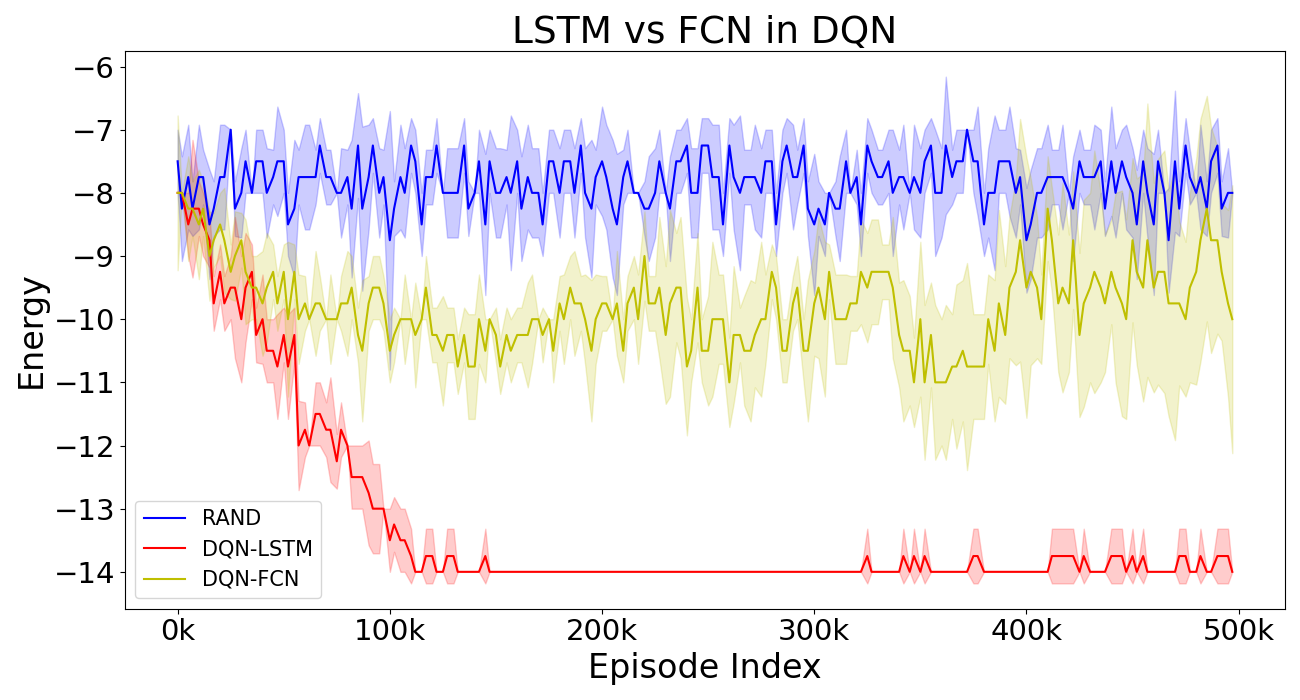}
\caption{\label{fig:lstm_vs_fcn}
    Comparison of DQN with Fully-Connected Network (FCN) vs with LSTM.
    Learning curves of RAND (blue),
    DQN with FCN (yellow),
    and DQN with LSTM (red)
    on the 500K-episode-trial of 36mer.
    X-axis is the index of episodes in progression.
    Y-axis denotes the energy found by the agent.
    The curves are plotted with the moving minimum of 200 episodes.
    The shaded area represents the standard deviation.
    Experiments are repeated four times on four random seeds for both RAND and DQN methods.
}
\end{figure}

In Fig.~\ref{fig:lstm_vs_fcn},
we show the effect of different NN architectures on DQN performance.
DQN with Fully-Connected Network (FCN) is used as a control.
The FCN architecture has the same number of layers and roughly the same number of trainable parameters as in the LSTM setting.
As seen from Fig.~\ref{fig:lstm_vs_fcn} yellow curve,
DQN with FCN is much less stable than with LSTM,
and displays ``catastrophic forgetting” during learning~\cite{geron2019hands}, the result can get worse and unstable with more episodes.

Our work demonstrates that LSTM can endow DQN with improved learning capability.
Intuitively, the LSTM captures long range interactions which is more suitable for both SAW and protein folding processes.
We hypothesize that there are ``motif''-like features learned by LSTM,
and our sequential one-hot encoded state representations capture the long-term dependencies in the time-series HP folding data.

LSTM has also been shown to be a good choice for the NN architecture in other DRL applications.
In 2019, OpenAI Five~\cite{openAIFIVE_berner2019dota} made headlines by defeating professional e-sports DOTA2 human teams, making use of a single-layer LSTM in their NN design.
More recently in 2021, EfficientZero~\cite{efficientZero_2021} shows for the first time a DRL algorithm to achieve super-human performance on Atari games with very limited data.
EfficientZero is able to match DQN's performance at 200 million frames while using 500 times less data using a LSTM-based NN architecture.
Particular to physics,
in 2022, the application of DRL for magnetic control of tokamak plasmas in nuclear fusion~\cite{Nature_2022_DRL_nuclearPlant} utilized an LSTM core in their network,
a design choice which the authors made after comparing with the FCN architecture.

\subsection{Sample Efficiency}

In our experiments, we conduct 100K SAW episodes per run for both 20mer-A and 20mer-B.
Existing exhaustive enumeration result~\cite{LUND_irback2002enumerating}
shows there are 41889578 ($\sim 4 \times 10^7$) complete SAW for 20mer (not including the incomplete SAW).
As a rough conservative estimate, a 100K sample size  makes up  less than $0.2\%$ of the 20mer SAW sample space.

For a 24mer, we use 500K SAW episodes per run.
The total number of possible complete SAW for 24mer is 2158326727
($\sim2 \times 10^9$) complete SAW (not including the incomplete SAW).
Our 500K sample size only constitutes less than $0.02\%$ of 24mer SAW sample space. Similarly for 36mer, we estimate that the 500K sample size is less than $2.5 \times 10^{-7}\%$ of all its SAW.

For our DRL experiments on benchmark sequences,
only a very tiny portion of the possible SAW search space gets sampled.
Yet we are able to reach best-known energies for all selected benchmark sequences.
We believe this shows that
DRL with LSTM is able to be sample efficient for medium to larger $N$.

\subsection{Search Effectiveness}

To investigate whether our DRL method would benefit from additional interventions,
we conduct control experiments with and without two manual pruning heuristics.
The first heuristic is to avoid having half of the actions to be consecutive `Forward',
so as to prevent the formation of elongated rod chains which are bad for generating HP contacts.
The second pruning heuristic is to early-stop a futile SAW if the remaining unprocessed HP units cannot yield a better solution.
These two pruning heuristics are designed to reduce the size of the search space with the help of human domain knowledge about the HP model.

\begin{figure}
\centering
\includegraphics[width=90mm]{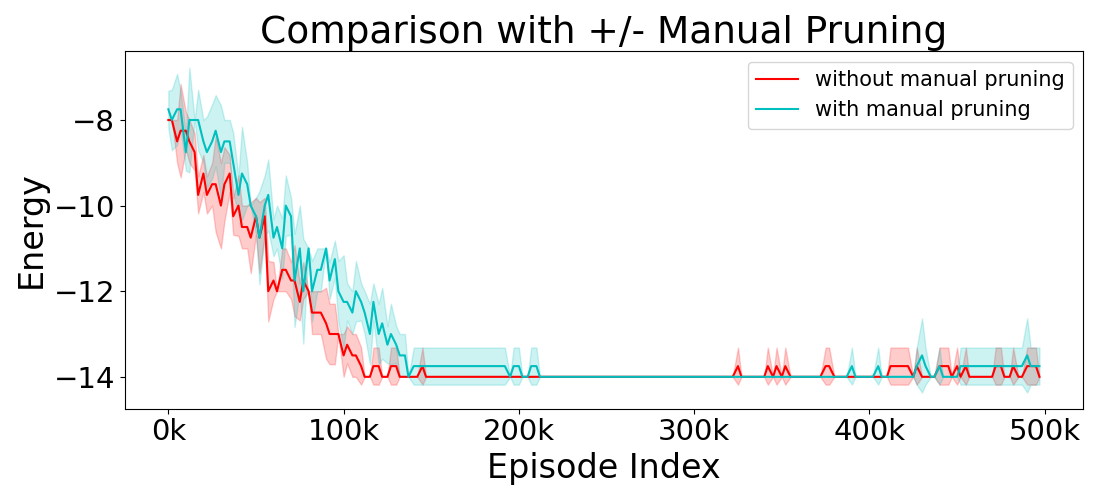}%
\caption{\label{fig:control_manualPrune} Default DRL algorithm (red) versus DRL with additional manual pruning heuristics (cyan)
on the 500K-episode-trial of 36mer.
    X-axis is the index of episodes in progression.
    Y-axis denotes the energy found by the agent.
    The curves are plotted with the moving minimum of 200 episodes.
    The shaded area represents the standard deviation.
    Experiments are repeated four times on four random seeds for both conditions.
    }
\end{figure}

Fig.~\ref{fig:control_manualPrune} shows default DRL (without manual pruning heuristics) versus addition of the manual pruning heuristics.
While the intention of manual pruning is to guide the DRL at searching in more relevant and deserving space,
the default DRL algorithm with the intrinsic value approximation can already traverse the search space more effectively on its own,
as shown by the faster convergence to best-known solutions.

In Chapter 8 of~\cite{RLsuttonBook_sutton2018reinforcement},
Sutton and Barto discuss how RL can learn to focus the computation on ``relevant states" resulting in more effective search process.
It is possible for RL to find an optimal solution by visiting only a small fraction of the states, even without having to visit some ``irrelevant states" at all.
The Fig.~\ref{fig:control_manualPrune} comparison with and without manual pruning of our DRL method
shows that the algorithm is able to learn an effective and intelligent search process without the need for heuristics even if they seem reasonable.

\begin{figure}
\centering
\includegraphics[width=66mm]{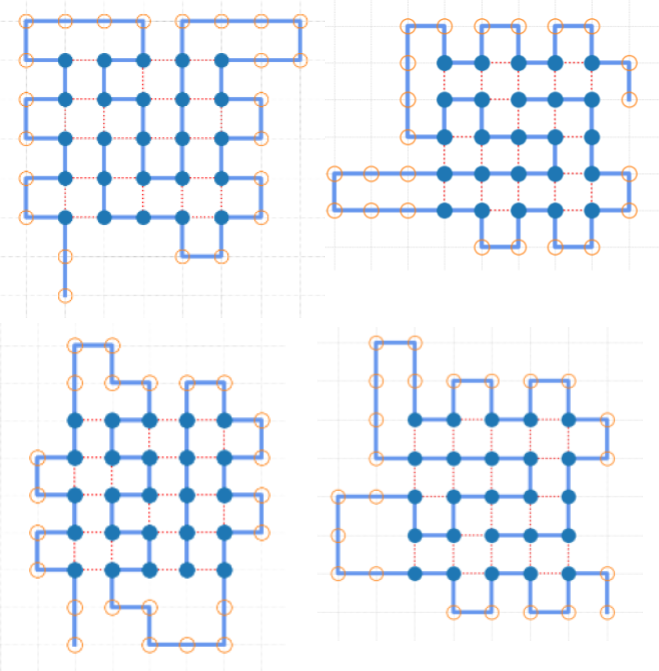}%
\caption{\label{fig:distinctSol48mer} Examples of four best-known solutions found by DRL agent for 48mer with $E_{\text{state}}=-23$ (to accompany Table~\ref{tab:intelli_search}).
Note these solutions are invariant to translational and rotational symmetries.
}
\end{figure}

Another aspect of the intelligent search from our DRL algorithm is
the capability to escape local optima and find many distinct best-known solutions during the search,
as shown in Table~\ref{tab:intelli_search}.
\begin{table}
\caption{\label{tab:intelli_search}%
Number of distinct best-known solutions
$E_{\text{state}}=-23$,
and next best solutions
$E_{\text{state}}=-22$
from the DRL experiments for 48mer, over four repeated trials with different random seeds.
Note the count of distinct solutions are on a \textbf{per-trial} basis, i.e.,
there can be overlap of solutions among trials.
}
\begin{threeparttable}
        \centering
        \resizebox{.88\linewidth}{!}{
\begin{tabular}{lccc}
\hline
\textrm{$N$-mer ID}&
\textrm{Trial}&
\textrm{\#Distinct Best-known Solutions\tnote{*}}&
\textrm{\#Distinct Next Best Solutions}\\
\hline
\multirow{4}{*}{48mer} & a & 48 & 91\\
& b & 12 & 17\\
& c & 24 & 80\\
& d & 10 & 60\\
\hline
\end{tabular}
}
        \begin{tablenotes}
        \footnotesize
            \item[*] Some examples of distinct best-known solutions for 48mer are shown in Fig.~\ref{fig:distinctSol48mer}.
        \end{tablenotes}
     \end{threeparttable}
\end{table}
We have already shown that the DRL agent can quickly converge to best-known solutions,
and in the exploitation phase of the learning curve,
the agent is able to keeps finding the best-known energy solutions.
We ask the following question---does the DRL agent memorize the previously encountered good solutions and simply regurgitate the memorized single encounter?
Table~\ref{tab:intelli_search} shows the number of distinct best-known solutions for the 48mer trials.
For all four trials,
the DRL agent is able to find many distinct best-known solutions.
This indicates that our DRL method does not get stuck in a local optimum,
but actively traverses the search space for better solutions.
The $\epsilon$-greedy strategy ensures the agent does continual exploration.
We present some of the best-known solutions found by our method for 
the 48mer in
Fig.~\ref{fig:distinctSol48mer}.

\subsection{Reproducibility and Reporting Standards}

Lastly, we comment on the standardization of experimental reporting.
Increasingly in the DRL research community, there is concern about reporting procedure and reproducibility issues~\cite{Limitation_henderson2018deep}.
RL is a trial-and-error learning distinguished by its exploration and exploitation trade-off.
The effect of randomness and variability can affect the results
of DRL experiments.

An influential paper that calls for a higher reporting standard to facilitate reproducibility~\cite{Limitation_henderson2018deep} recommends the following procedure for DRL experiments: 1) Hyper-parameter configurations should be provided; and
2) Multiple trials with different random seeds should be run for reporting.

Unfortunately,
few of the related RL work we compared with
follow the above guidelines.
Most did not list a working set of hyper-parameters, especially the number of time-steps or episodes they use in their trials.
Furthermore,
we do not know how many trials were attempted.

Without knowing the number of trials and the scale of their iterations,
we cannot have a fair comparison or meaningfully critique other reported RL results.
In our paper,
we document the hyper-parameters used, number of episodes and trials.
We welcome future fair comparisons with our DRL design and implementation.

We also notice that past literature often only report the energy values obtained by their proposed algorithms without providing additional important evidence like the folding and conformations.
Apart from the lack of details with such reporting,
this also gives a false impression that the benchmark HP sequences have unique optimal conformations.
Few studies have pointed out many of the benchmark sequences are in fact highly degenerate.
For example, we found that the 36mer can have more than $180$ distinct conformations that all have the best-known energy of $-14$.

As a contribution to the existing popular benchmark,
we release a conformation database
of the benchmark sequences as a Zenodo open data repository (link in our GitHub repo in Data Availability Statement).
As far as we are aware,
this is the first conformation database to show the best-known solutions for benchmark HP sequences,
plus detailing all the possible distinct best-known solutions found.
A preview of the conformation database is shown in~\ref{app:conf_DB}.
We believe such statistical characterization (e.g. the count of degenerate best-known and next best solutions) of benchmark HP sequences are of considerable interest not only for energy landscape profiling,
but also to traditional greedy algorithms such as the classical Monte Carlo methods when designing and tuning parameters.
Biophysical researchers may also benefit from analyzing the differences and patterns among the multiple conformations of best-known or next best energies to gain insight on structure prediction.

\section{Conclusion}

We demonstrate the effectiveness of applying DRL to the HP model for protein folding.
Our DRL setup achieves conformations of best-known energies in benchmark HP sequences ranging from length 20 to 50,
which is better than previous results in the literature.
The LSTM architecture endows DRL with enhanced learning capacity,
as LSTM's sequential representation ability captures long range interactions, which are key to protein folding.
The search process of DRL can traverse relevant state space effectively without the use of
manual pruning heuristics.
It can also find multiple distinct best-known solutions per trial.
To encourage a higher reporting standard and help with reproducibility,
we list our working configurations and hyper-parameters.


For future work, we can investigate recent developments on model-based DRL methods such as the previously mentioned EfficientZero~\cite{efficientZero_2021},
which uses Monte Carlo Tree Search with multiple step look-ahead to help improve performance.

Another interesting direction to explore is to extend the optimal SAW path finding task to other variants of the HP model,
such as the HPNX model, which accounts for the neutral and charged amino acid residues besides H and P ones~\cite{HPNX_1999_backofen}.

Lastly, we note that although the HP model is one of the most extensively studied protein models~\cite{hart2006proteinMIT, istrail2009combinatorial},
its simple model is not designed for modelling practical protein folding.
Interested readers can refer to a review on the assumptions of the HP model from Ken Dill et al.~\cite{globular_dill1995principles}.

Our DRL ``recipe" follows the basic DQN structure from the DeepMind paper~\cite{mnih2015human},
and we show that the prototypical DQN algorithm is sufficient to be applied to the HP model protein folding but key ``ingredients" are needed.
Unlike the previous works' attempt of using DQN (or advanced variants of DQN),
we believe the `secret sauce' is a suitable state representation and NN architecture.
We present our considerations and design choices in this paper as a possible prototype for future RL application to HP model research,
which should be extensible to incorporate more recent DRL developments.

Here we have presented not only a working framework viewing the HP model as a SAW path-finding optimization task,
but a DRL solution that is state-of-the-art.
We hope this will help both the HP model and RL communities with continued progress and reproducible results.

\section*{Acknowledgments}
We thank Ken Sung, Wong Limsoon, Tay Yong Chiang from the NUS Department of Computer Science for helpful comments and discussions. This research is financially supported by the National Natural Science Foundation of China (Project No. 21973080), the Research Grants Council of Hong Kong (Project No. 21302520), and City University of Hong Kong (Project No. 7005601).

\section*{Data Availability Statement}

\textbf{Source code} available as a public open-source repository at \href{https://github.com/CompSoftMatterBiophysics-CityU-HK/Applying-DRL-to-HP-Model-for-Protein-Structure-Prediction}{GitHub Repo URL}:
\url{https://github.com/CompSoftMatterBiophysics-CityU-HK/Applying-DRL-to-HP-Model-for-Protein-Structure-Prediction}.

\textbf{Conformation database} showing the distinct conformations of best-known and next best energies is available as a Zenodo open data repository (maintained by CERN) via a link in our GitHub repository.
As a preview, the complete \textbf{conformation action sequences and the corresponding conformation images for 48mer's}
best-known and next best solutions are available in
the supplementary material.

\appendix

\section{Conformation Database}
\label{app:conf_DB}

\begin{figure}
\centering
\includegraphics[width=113mm]{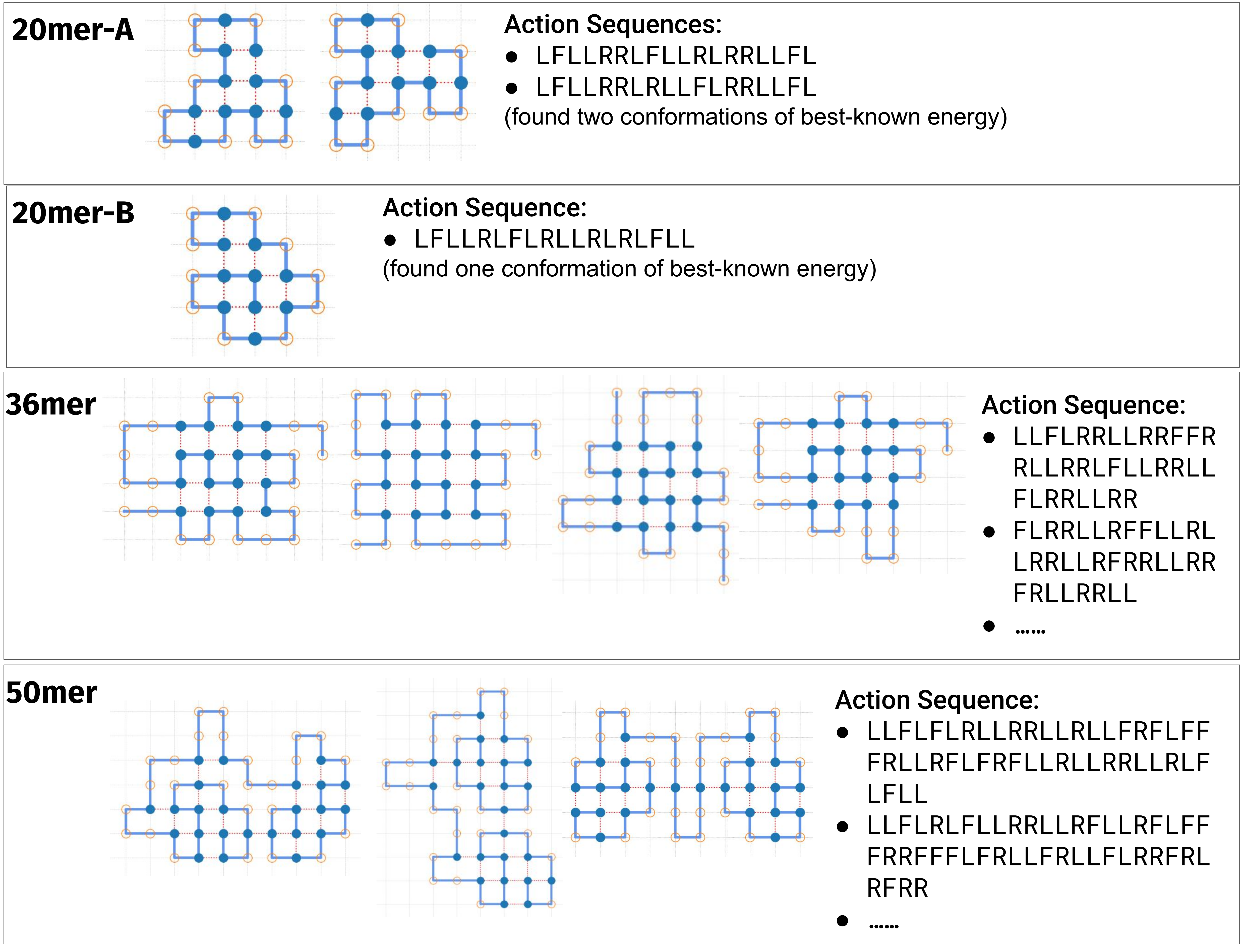}%
\caption{\label{fig:selected_degenerate_best_known} Examples of best-known solutions found by DRL agent
along with the action sequences stored in the conformation database.
}
\end{figure}

The conformation database is to accompany all the sequences in Table~\ref{tab:table_istrail_benchmark_info} from the Istrail benchmark.
We will provide the count, action sequences,
and folding conformation images for distinct best-known
and next best
solutions found by DRL agent,
similar to the effect of combining Table.~\ref{tab:intelli_search}
and Fig.~\ref{fig:distinctSol48mer}.
Some examples of conformations of best-known energies and the corresponding action sequences from our conformation database are shown
in Fig.~\ref{fig:selected_degenerate_best_known}.
Please refer to the \textbf{Supplementary Material}
for all the conformation action sequences and the corresponding conformation images for 48mer's
best-known and next best solutions (from Table.~\ref{tab:intelli_search}).
A preview of the supplementary material is given in Fig.~\ref{fig:supp_48mer}.
\begin{figure}
\centering
\includegraphics[width=113mm]{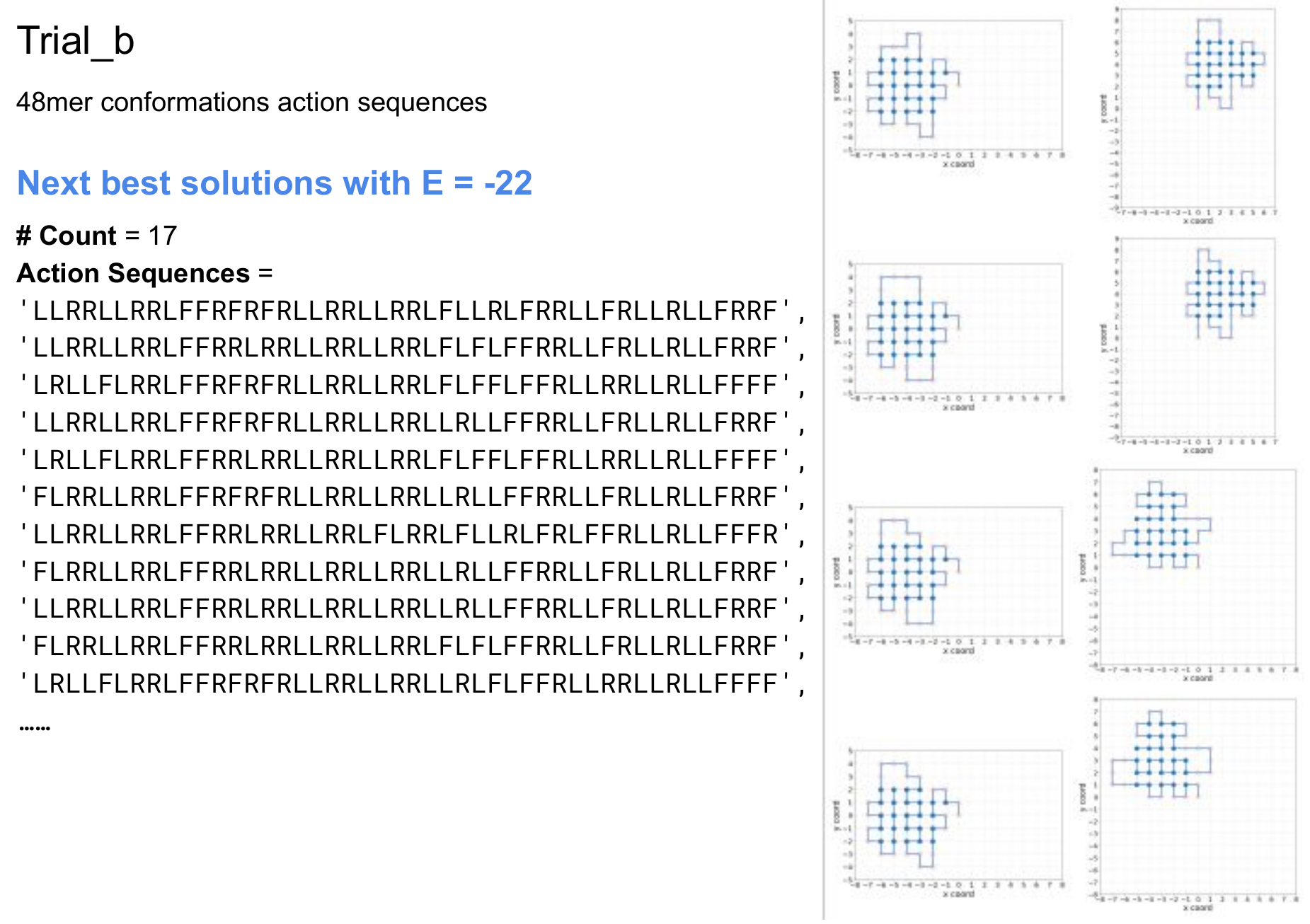}%
\caption{\label{fig:supp_48mer} Screenshots of conformations of next best energies and their action sequences from Trial-b (from Table.~\ref{tab:intelli_search}) of 48mer as provided in the Supplementary Material.
}
\end{figure}
We believe this database will help the HP model research community with additional insight on the popular benchmark sequences and in greedy algorithm designs.

 \bibliographystyle{elsarticle-num} 
 \bibliography{cas-refs}





\end{document}